\pdfoutput=1

\documentclass[11pt]{article}

\usepackage[]{acl}

\usepackage{times}
\usepackage{latexsym}

\usepackage[T1]{fontenc}

\usepackage[utf8]{inputenc}

\usepackage{microtype}
\usepackage{graphicx}
\usepackage{amsmath}
\usepackage{booktabs}
\usepackage{multirow}
\usepackage{multicol}
\usepackage{enumitem}

\newcommand{\converser}{\textsc{Converser}}

\title{\converser: Few-Shot Conversational Dense Retrieval \\ with Synthetic Data Generation}

\author{Chao-Wei Huang$^{\dag\ddag}$\quad Chen-Yu Hsu$^\dag$\quad Tsu-Yuan Hsu$^\dag$\Thanks{ Equal contribution} \quad Chen-An Li$^\dag$\footnotemark[1] \quad Yun-Nung Chen$^\dag$\\
  $^{\dag}$National Taiwan University, Taipei, Taiwan\\
  $^{\ddag}$Taiwan AI Labs, Taipei, Taiwan \\
  \texttt{f07922069@csie.ntu.edu.tw\quad y.v.chen@ieee.org} \\}

\begin{document}
\maketitle
\begin{abstract}
Conversational search provides a natural interface for information retrieval (IR).
Recent approaches have demonstrated promising results in applying dense retrieval to conversational IR.
However, training dense retrievers requires large amounts of in-domain paired data.
This hinders the development of conversational dense retrievers, as abundant in-domain conversations are expensive to collect.
In this paper, we propose \converser, a framework for training conversational dense retrievers with at most 6 examples of in-domain dialogues.
Specifically, we utilize the in-context learning capability of large language models to generate conversational queries given a passage in the retrieval corpus.
Experimental results on conversational retrieval benchmarks OR-QuAC and TREC CAsT 19 show that the proposed \converser~achieves comparable performance to fully-supervised models, demonstrating the effectiveness of our proposed framework in few-shot conversational dense retrieval.\footnote{All source code and generated datasets are available: \url{https://github.com/MiuLab/CONVERSER}}
\end{abstract}

\section{Introduction}
Conversational information retrieval (CIR) has been an important area of research in recent years, aiming to retrieve relevant information from a large corpus of text in a conversational format.
It has gained considerable interest due to its potential to deliver information in a natural format in response to a user's queries.
Unlike traditional IR, CIR poses distinctive challenges, including its multi-turn and context-dependent nature, which require more nuanced approaches~\cite{yu2021few,fang-etal-2022-open}.

Dense retrieval methods have demonstrated their ability to understand the semantics of complex user queries and shown promising performance on open-domain retrieval~\cite{karpukhin-etal-2020-dense}.
One of the major obstacles to conversational dense retrieval is the scarcity of training data, given the high cost and extensive time to collect high-quality information-seeking conversations~\cite{adlakha-etal-2022-topiocqa}.
Previous work has explored various approaches to address this issue~\cite{pmlr-v162-dai22a,kim-etal-2022-generating}.
However, most methods still rely on the assumption that a large amount of in-domain data is present and build data augmentation models upon it.

In this paper, we aim to develop a few-shot conversational dense retrieval model that can effectively retrieve relevant passages based on a small number of in-domain dialogues.
To achieve this, we leverage the in-context learning capability of large language models (LLMs) to generate synthetic passage-dialogue pairs with few-shot demonstrations.
Specifically, in-domain passages are sampled from the retrieval corpus, and dialogues are synthesized by asking LLMs to generate a series of queries based on a few examples.
We also employ a self-consistency filtering mechanism to automatically discard inconsistent generated queries, ensuring the accuracy and reliability of the generations.

We conduct experiments on two benchmark datasets, including OR-QuAC~\cite{qu2020open} and TREC CAsT 19~\cite{Dalton2019Cast}.
The experimental results demonstrate that our proposed framework, \converser, performs comparably to fully-supervised models that are trained on \emph{thousands} of annotated dialogues while using only 6 examples at most.
Furthermore, analyses show that \converser~rivals other data augmentation methods that utilize full in-domain datasets, demonstrating its effectiveness.

\section{Related Work}

\paragraph{Conversational Dense Retrieval}
Conversational dense retrieval poses a unique challenge in that the questions are context-dependent.
Prior works have explored various modeling techniques for conversational history to address this challenge~\cite{huang2018flowqa,choi-etal-2018-quac,yeh-chen-2019-flowdelta,chiang2020empirical}.
However, these works only examined the modeling ability for conversational question answering (CQA), where the relevant passages are provided.

More recently, \citet{qu2020open} proposed OR-ConvQA, which extends CQA to the open-domain setting where a retrieval module is required.
ConvDR~\cite{yu2021few} utilizes an ad-hoc dense retriever and manually rewritten context-independent queries for training few-shot retrievers and rerankers, while our method does not require an ad-hoc model and additional annotation.
Others have explored various methods for encoding conversational queries~\cite{li2021graph,fang-etal-2022-open,wu-etal-2022-conqrr,liang2022multifaceted}, which are orthogonal to our work.

\subsection{Synthetic Data Generation for Dense Retrieval}
Due to the data-hungry nature of dense retrievers, synthetic data generation for dense retrieval has drawn considerable interest.

Previous works have worked on generating information-seeking conversations via transforming documents~\cite{pmlr-v162-dai22a,kim-etal-2022-generating} or web search sessions~\cite{mao-etal-2022-convtrans}.
However, these methods all require training query generators with conversational data, which does not mitigate the data scarcity issue.
Our method requires only 6 in-domain dialogues with their relevant passages and demonstrates comparable performance to models trained on thousands of manually annotated dialogues.

InPars~\cite{10.1145/3477495.3531863} and Promptagator~\cite{dai2023promptagator} are the most closely related works to our method.
They both proposed to generate synthetic queries with LLMs from few-shot examples, which achieved comparable performance to supervised methods in dense retrieval.
Inspired by these works, our method further extends few-shot query generation to the conversational setting.
We propose novel techniques for generating conversational queries and show that they are crucial to handle the unique challenges of conversational dense retrieval. 

\begin{figure*}[!t]
    \centering
    \includegraphics[width=\linewidth]{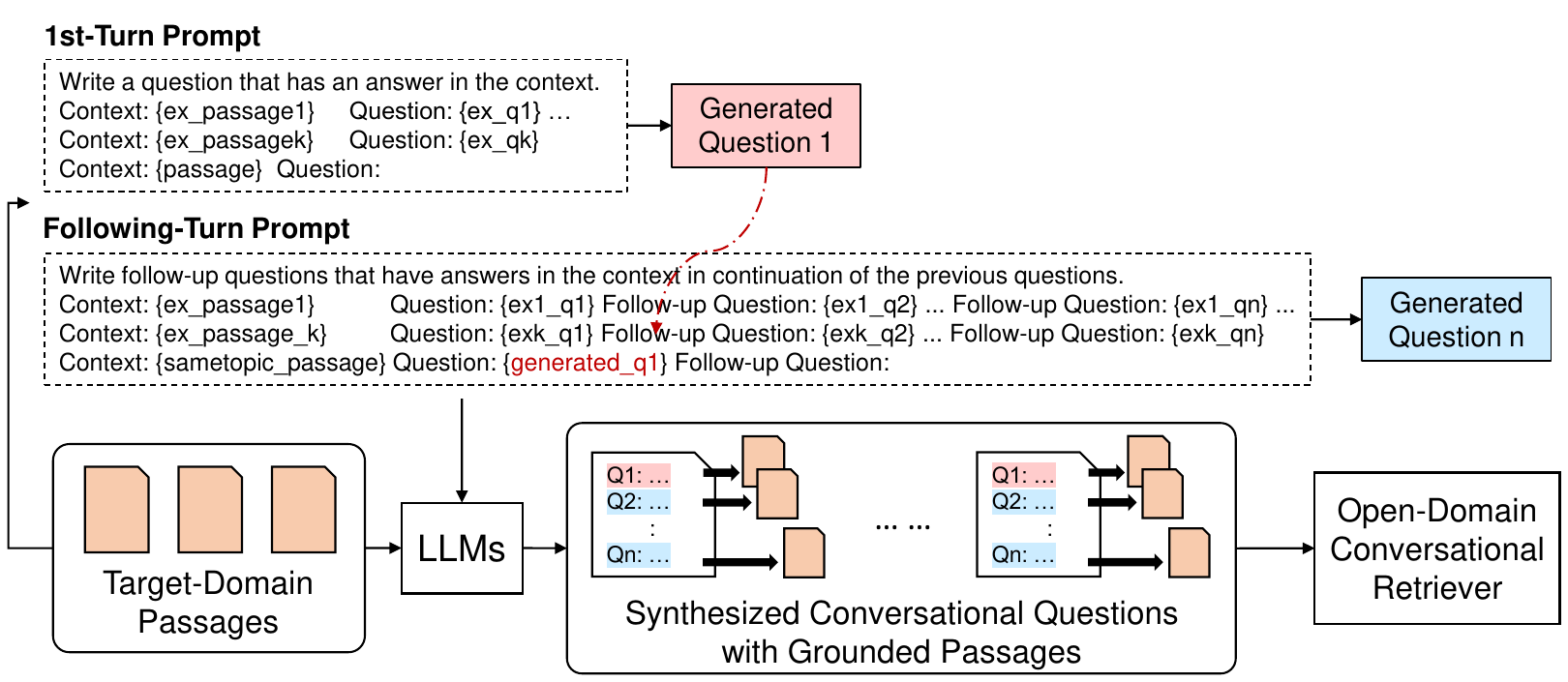}
    \caption{Illustration of our proposed framework.}
    \label{fig:framework}
\end{figure*}

\section{Proposed Method: \converser}
We propose few-shot \underline{convers}ational d\underline{e}nse \underline{r}etrieval with synthetic data generation, \converser, which aims to generate synthetic conversational queries given few examples.
More formally, given a conversational retrieval task $T$, its retrieval corpus $\mathcal{P}_T$, and $k$ examples, we aim to generate synthetic conversational query-passage pairs $\{\hat{C}_1, \cdots, \hat{C}_n\}$ for training dense retrievers.

\subsection{Few-Shot Conversational Query Generation}
The core of our method is \textit{few-shot query generation}.
We leverage the in-context learning ability of LLMs~\cite{brown2020language} to generate conversational queries.
Specifically, we start with $k$ examples $\{C_1, C_2, \cdots, C_k\}$, where each $C_i$ is a conversation represented as a series of query-passage pairs, $(q_i^1, p_i^1), \cdots, (q_i^{n_i}, p_i^{n_i})$, with $n_i$ denoting the length of $C_i$.
Using these examples, we construct the following template $\mathcal{T}$ as a few-shot demonstration for LLMs:
\begin{equation*}
    \left[ (p_1^{n_1}, q_1^1, \cdots, q_1^{n_1}), \cdots, (p_k^{n_k}, q_k^1, \cdots, q_k^{n_k}) \right]
\end{equation*}
Note that we always choose the relevant passage that corresponds to the last query in the examplar, indicating that the last query $q_i^{n_i}$ is generated given $p_i^{n_i}$ and previous queries $q_i^1, \cdots, q_i^{n_i-1}$.

The generation process for a synthetic conversation starts with randomly sampling a passage $\hat{p}$ from the retrieval corpus, i.e., $\hat{p} \sim \mathcal{P}_T$.
We concatenate the template and the sampled passage to form an input text sequence $\left[ \mathcal{T}, \hat{p} \right]$.
An LLM is employed for generating synthetic queries.
It is expected to generate the first query $\hat{q}_1$ that is relevant to $\hat{p}$ based on the provided examples.
We then append $\hat{q}_1$ to the input sequence, forming the input sequence for generating the next query $\hat{q}_2$, and so forth.
We sequentially perform the generations for a conversation until a predefined number of turns is reached.

\subsection{Two-Stage Generation}
One unique characteristic of conversational queries is that the queries are \textit{context-dependent}~\cite{choi-etal-2018-quac} except for the first query, which should be a self-contained query without any ambiguity.
To address this difference, we propose to split the generations into two-stage: first query generation and follow-up query generation.
When generating the first query for each conversation, we use an alternative template $\mathcal{T}_1 = \left[ p_1^{1}, q_1^1, \cdots, p_k^{1}, q_k^1 \right]$, which contains only the first queries and their relevant passages of the examples.
We then replace $\mathcal{T}_1$ with $\mathcal{T}$ for generating all the follow-up queries.
In practice, we found that this two-stage approach reduces the number of generated first queries that are not self-contained and thus ambiguous.

\subsection{Passage Switching}
In a conversation, relevant passages may vary for different queries.
To this end, we incorporate passage switching into the generation process.
We randomly replace the current passage $\hat{p}$ with a related passage $\hat{p}^{\prime}$ in each turn with a probability $p_{ps}$.
The LLM is expected to generate queries based on the new passage.

\subsection{Consistency Filtering}
The generation process sometimes generates queries that are nonsensical, degenerated, ambiguous, or not grounded by the given passage.
We adopt a filtering mechanism via ensuring \textit{round-trip consistency}~\cite{alberti-etal-2019-synthetic}.
We follow the procedure in~\citet{dai2023promptagator}, where an initial retriever is trained on all synthetic query-passage pairs.
For each synthetic pair $(\hat{q}, \hat{p})$, we use the initial retriever to retrieve the most relevant passages for $\hat{q}$ from $\mathcal{P}_T$.
We keep the pair $(\hat{q}, \hat{p})$ only if $\hat{p}$ is in the top-k retrieved passages.

\begin{table*}[!ht]
\centering 
\begin{tabular}{l|ccc|cc}
\toprule
\multicolumn{1}{c|}{\multirow{2}{*}{\bf Method}} & \multicolumn{3}{c|}{\bf OR-QuAC} & \multicolumn{2}{c}{\bf CAsT-19} \\ 
& \bf MRR@5 & \bf R@5 & \bf MAP@10 & \bf MRR & \bf NDCG@3 \\
\midrule 
Supervised OR-ConvQA~\citep{qu2020open} & 22.5 & 31.4 & - & - & - \\
Supervised DPR & 50.5 & 64.7 & 49.7 & 29.4 & 19.1 \\
Few-Shot \converser~(Ours) & 49.6 & 63.4 & 48.7 & 35.8 & 21.4 \\
\bottomrule
\end{tabular}
\caption{Evaluation results (\%). We report the result of OR-ConvQA from the original paper.}
\label{tab:results}
\end{table*}

\begin{table}[!ht]
\centering 
\begin{tabular}{l|cc}
\toprule
\multicolumn{1}{c|}{\multirow{2}{*}{\bf Method}} & \multicolumn{2}{c}{\bf OR-QuAC} \\ 
& \bf MRR@5 & \bf R@5 \\
\midrule
OR-QuAC & 50.5 & 64.7 \\
WikiDialog (31k) & 44.6 & 58.2 \\
\midrule
\converser~(31k) & 46.8 & 61.5 \\
- Two-Stage & 45.1 & 59.9 \\
- Consistency Filtering & 45.2 & 59.8 \\
- Passage Switching & 45.6 & 58.1 \\
- Only 1-Shot & 42.1 & 55.2 \\
\bottomrule
\end{tabular}
\caption{Results of ablation study. We use the identical training procedure and training data size for each experiment to make them comparable.}
\label{tab:ablation}
\end{table}

\section{Experiments}
To evaluate if our generated conversational questions can help train a conversational retriever, we conduct experiments on a conversational question answering dataset, OR-QuAC~\cite{qu2020open}, and a conversational search benchmark, TREC CAsT-19~\cite{Dalton2019Cast}.

\subsection{Experimental Setup}
We describe our experimental setup in the section.
Additional details can be found in Appendix~\ref{appendix:details}.

\paragraph{Few-Shot Examples}
We manually select 6 examples for OR-QuAC and 5 examples for CAsT-19 and use the same set of examples in all experiments.
Due to resource constraints, we use the remaining 15 conversations for evaluating on CAsT-19 without performing 5-fold cross-validation.

\paragraph{Generation}
We employ LLaMA-13B~\cite{touvron2023llama} as our pretrained LLM, which is not instruction-tuned and is open to the research community.
We use nucleus sampling~\cite{Holtzman2020The} for decoding and set $\text{top\_p} = 0.95$, $\text{temperature} = 0.75$.
We generate 427k turns (61k conversations) for OR-QuAC and 230k turns (32k conversations) for 
An example of generation results can be found in Section~\ref{sec:example}.

\paragraph{Retrieval Corpus}
We generate synthetic conversations based on the retrieval corpus for each task respectively.
For OR-QuAC, we use the provided 11M passages from English Wikipedia.
For TREC CAsT-19, we use the official passage collection, which consists of 8M webpage passages from MSMARCO~\cite{bajaj2016ms} and 30M Wikipedia passages from TREC-CAR~\cite{dietz2017trec}.

\paragraph{Model Details}
We follow the procedures from DPR~\cite{karpukhin-etal-2020-dense} to train our retrievers and use BERT-base as the pretrained model.
We concatenate all previous queries and the current query as the input to the retriever.
Additional details can be found in Appendix~\ref{appendix:details}.

\paragraph{Baseline Systems}
\begin{itemize}[align=parleft,left=0pt..1em,itemsep=0pt]
    \item \textbf{OR-ConvQA}: A supervised dense retriever trained on OR-QuAC~\cite{qu2020open}.
    \item \textbf{DPR}: We train a DPR model~\cite{karpukhin-etal-2020-dense} on the training set of OR-QuAC for a fair comparison.
\end{itemize}

\subsection{Main Results}
Table~\ref{tab:results} shows the experimental results.
Note that both ConvDR and WikiDialog utilized multiple additional datasets and techniques, which are complementary to our method.
On the OR-QuAC dataset, our proposed \converser~outperforms the supervised baseline OR-ConvQA by a large margin and performs comparably to the supervised DPR trained on OR-QuAC.
This result demonstrates the effectiveness of our few-shot generation strategy, as our model trained on a synthetic dataset based on only 6 annotated examples can rival the performance of supervised DPR, which is trained on 4000 annotated dialogues.

On CAsT-19, \converser~outperforms supervised DPR, which is trained on OR-QuAC.
This shows that our task-specific generation strategy can effectively synthesize conversational queries on a new task given a few examples of the new task.
Our proposed method provides better adaptability without requiring another supervised dataset as done in conventional transfer learning.

\subsection{Ablation and Comparative Study}
We conduct an ablation study on different settings of our proposed method, where we remove one component at a time to validate its effectiveness.
We also compare our method with two datasets: OR-QuAC and WikiDialog~\cite{pmlr-v162-dai22a}.
To ensure the results are comparable, we limit the size of every dataset to 31k turns, which is the same as the training set of OR-QuAC.
The training process and hyperparameters are also identical for all datasets.
For WikiDialog, we subsample the original WikiDialog dataset and use it to fine-tune a retriever, without further fine-tuning on OR-QuAC.
The results are shown in Table~\ref{tab:ablation}.

Given the same number of synthesized turns, our \converser~outperforms WikiDialog, which requires supervised conversational datasets for training a query generator.
This result validates the effectiveness of our proposed few-shot generation method.
The ablation study demonstrates that all of our proposed components contribute to the improvement.

\begin{table*}[t!]
\centering
\small
\begin{tabular}{lp{13cm}}
\toprule
\multicolumn{2}{l}{\textbf{Generated Question Example}} \\
\midrule
\textbf{Title} & Palazzo Primoli \\
\textbf{Passage} & Palazzo Primoli Palazzo Primoli is a Palazzo in Rome, Italy. It is owned by the city of Rome and houses several museums and collections. The palazzo was built in the seventeenth century. In 1901 Count Giuseppe Primoli (1851–1927) became its sole owner. He extended and partly modernised the palazzo with a new facade and entrance between 1901 and 1911. The Count's maternal grandparents were Charles Lucien Bonaparte and Zénaïde Bonaparte, and the Count brought together a collection of objects (now the Museo Napoleonico), documenting the relationship between Rome and the Bonaparte family. He also was an avid photographer. In 1927 Giuseppe Primoli donated the palazzo and his collections to the municipality of Rome. The Museo Napoleonico is located on the palazzo's ground floor, and the third floor is occupied by the Museo Mario Praz, the former residence of Mario Praz. Also located in the palazzo are the Count's library and photographic archive. \\
\midrule
\textbf{WikiDialog} & Q1: What is known about Palazzo Primoli? \\
& Q2: Are there any other interesting aspects about this article? \\
& Q3: When was it built? \\
& Q4: Who were the previous owners of the palazzo? \\
& Q5: What was his role in the palazzo? \\
& Q6: What is the most interesting aspect of this article? \\
\midrule
\textbf{\converser} & Q1: What is the name of the building that houses the Museo Napoleonico? \\
& Q2: What is the Palazzo Primoli best known for? \\
& Q3: What year was it donated? \\
& Q4: What famous collection was donated? \\
& Q5: What other collections are housed in this museum? \\
& Q6: Who was the Count's maternal grandparents? \\
\midrule
\textbf{\converser} & Q1: What is Palazzo Primoli in regards to Rome? \\
 only 1 example & Q2: What is the Palazzo Primoli's collection of objects? \\
& Q3: What is the Museo Napoleonico? \\
& Q4: What was the Count's library? \\
& Q5: What was the Count's photographic archive? \\
\bottomrule
\end{tabular}
\vspace{-1mm}
\caption{A qualitative example from WikiDialog, \converser, and \converser~with only 1 example.}
\label{tab:qualitative} 
\end{table*}

\subsection{Effect of Generated Data Size}

\begin{figure}[!t]
    \centering
    \includegraphics[width=\linewidth]{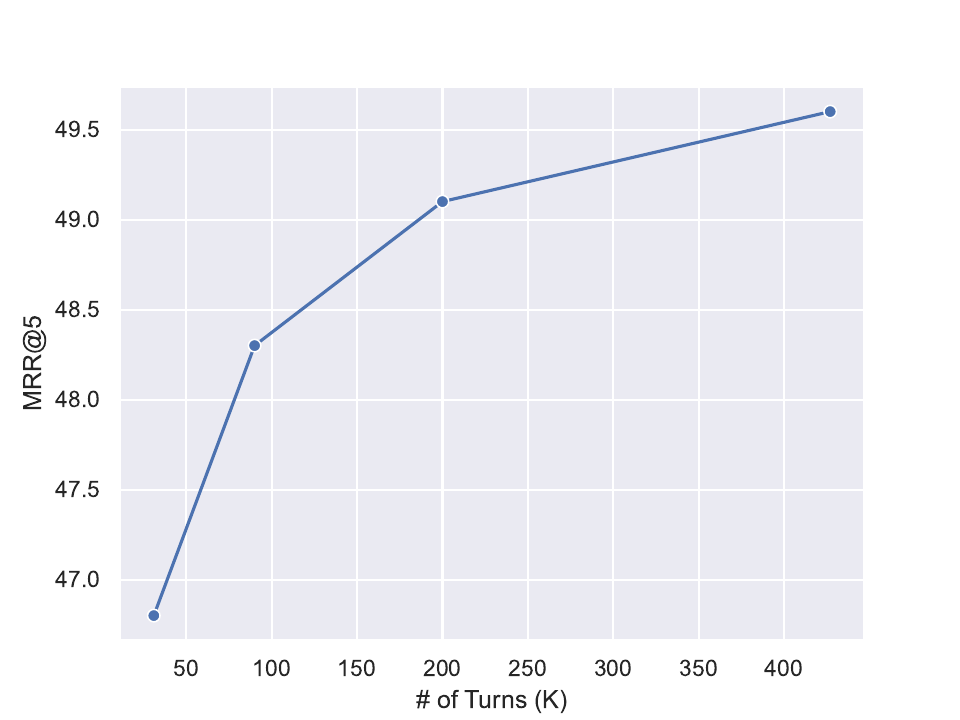}
    \caption{MRR@5 with regard to different number of generated turns on OR-QuAC.}
    \label{fig:data_size}
\end{figure}

We explore the impact of the generated data size on the performance, where we conduct a series of experiments, systematically varying the number of generated turns used for training presented in Figure~\ref{fig:data_size}.
It clearly illustrates that as the number of turns increases, the system's performance improves significantly. This finding highlights the crucial role of conversational data in enhancing the effectiveness of our model.

\section{Qualitative Study}
\label{sec:example}
We present a generated example in Table~\ref{tab:qualitative} to perform qualitative analysis.
WikiDialog is capable of generating follow-up questions.
However, it often generates generic queries, such as \textit{Are there any other interesting aspects about this article}.
On the other hand, \converser~with only 1 example suffers from a lack of diversity.
Due to limited demonstrations, it generates queries that are very similar to the only example it is given.
Our proposed \converser~can generate a context-independent first question and follow-up questions, demonstrating its effectiveness.

\section{Conclusion}
This paper introduces \converser, a synthetic data generation method for training few-shot conversational dense retrievers.
We leverage the in-context learning capability of LLMs and propose techniques that are designed for generating conversational queries.
Experimental results demonstrate that our proposed \converser~achieves comparable performance to fully-supervised models while only requiring 6 annotated examples.
Further analyses demonstrate that our method outperforms a fully-supervised data augmentation method.
Future work could explore instruction-following LLMs, better filtering mechanisms, and synthesizing specialized data for conversational dense retrieval, such as query rewrites.

\section*{Acknowledgements}
We thank the reviewers for their insightful comments. This work was financially supported by the National
Science and Technology Council (NSTC) in Taiwan, under Grants 111-2222-E-002-013-MY3, 111-2628-E-002-016, and
112-2223-E-002-012-MY5 and Google.

\bibliography{anthology,custom}
\bibliographystyle{acl_natbib}

\appendix

\section{Implementation Details}
\label{appendix:details}

\paragraph{Generation}
Text generation with language models often results in degeneration, i.e., repeating the same text sequence.
Hence, we heuristically filter out degenerated generations.
Initially, we examined the generation quality of LLaMA-7B.
However, it showed an increased amount of degeneration and queries of lower quality.
We have also tried several open-source instruction-tuned LLMs.
To our surprise, these models failed to generate conversational queries given instructions, with or without few-shot examples.
Using instruction-tuned LLMs for conversational query generation could be a direction for future exploration.
Generations are conducted on 2 NVIDIA V100 GPUs.
Generating one conversation takes roughly 10 seconds on a single GPU.

\paragraph{Training Details}
All retrievers are trained with a batch size of 64 queries.
We use in-batch negatives as it is found to be important~\cite{karpukhin-etal-2020-dense}.
We train all retrievers for 10 epochs with a learning rate of 2e-5.
To reduce GPU memory consumption, we use the DPR implementation with gradient cache~\cite{gao2021scaling}, enabling larger batch size.
The training process is done on 4 NVIDIA 2080Ti GPUs.

\paragraph{Evaluation Details}
We evaluate the models on the test sets of the evaluation datasets.
There are 20 conversations for evaluation in CAsT-19.
Previous work has conducted 5-fold cross-validation to address the lack of training in CAsT-19.
However, due to resource constraints, we could not run generations for 5 different sets of examples.
Hence, we manually select 5 conversations for building the few-shot examples and use the remaining 15 conversations for evaluation.

We report the most commonly-used evaluation metrics on each dataset: \textbf{MRR@5}, \textbf{R@5}, and \textbf{MAP@10} for OR-QuAC, and \textbf{MRR} and \textbf{NDCG@3} for CAsT-19.

\end{document}